\pdfoutput=1

\documentclass[11pt]{article}

\usepackage{ACL2023}

\usepackage{times}
\usepackage{latexsym}

\usepackage[T1]{fontenc}

\usepackage[utf8]{inputenc}

\usepackage{microtype}

\usepackage{inconsolata}

\usepackage{graphicx}               
\usepackage{tabularx}               
\usepackage{booktabs}
\usepackage{amsmath}
\usepackage{stmaryrd}
\usepackage{xcolor}
\usepackage{soul}

\newcolumntype{C}{>{\centering\arraybackslash}X}
\usepackage{multirow}               
\usepackage{diagbox}                
\usepackage{hhline}                 
\usepackage{color}                  
\usepackage{amsmath}                
\usepackage{amssymb}                
\usepackage{mathtools}              
\usepackage{enumitem}               

\usepackage{subfigure}
\usepackage{booktabs}
\usepackage{extarrows}
\usepackage{makecell}

\definecolor{RoseQuartzBg}{HTML}{F7CAC9}
\definecolor{RoseQuartz}{HTML}{F5A798}
\definecolor{Serenity}{HTML}{92A8D1}
\definecolor{OrangeRed}{rgb}{1.0, 0.27, 0.0}
\definecolor{Red}{rgb}{1.0, 0.0, 0.0}
\definecolor{Turquoise}{HTML}{0F4C81}
\usepackage{xparse}
\NewDocumentCommand{\lifu}{ mO{} }{\textcolor{OrangeRed}{\textsuperscript{\textit{Lifu}}\textsf{\textbf{\small[#1]}}}}
\NewDocumentCommand{\mo}{ mO{} }{\textcolor{blue}{\textsuperscript{\textit{Mo}}\textsf{\textbf{\small[#1]}}}}
\NewDocumentCommand{\minqian}{ mO{} }{\textcolor{teal}{\textsuperscript{\textit{Minqian}}\textsf{\textbf{\small[#1]}}}}
\NewDocumentCommand{\zhiyang}{ mO{} }{\textcolor{Red}{\textsuperscript{\textit{Zhiyang}}\textsf{\textbf{\small[#1]}}}}
\NewDocumentCommand{\sijia}{ mO{} }{\textcolor{Red}{\textsuperscript{\textit{Sijiia}}\textsf{\textbf{\small[#1]}}}}

\definecolor{fig_blue}{HTML}{023eff}
\definecolor{fig_green}{HTML}{6acc64}
\definecolor{fig_orange}{HTML}{ff7c00}
\definecolor{fig_red}{HTML}{e8000b}

\usepackage{graphics}

\title{Teamwork Is Not Always Good: An Empirical Study of Classifier Drift in Class-incremental Information Extraction}

\author{Minqian Liu, \ Lifu Huang
\\
Computer Science Department \\
  Virginia Tech \\ 
  {\tt \{minqianliu,lifuh\}@vt.edu}
  }

\begin{document}
\maketitle

\begin{abstract}

Class-incremental learning (CIL) aims to develop a learning system that can continually learn new classes from a data stream without forgetting previously learned classes. 
When learning classes incrementally, the classifier must be constantly updated to incorporate new classes, and the drift in decision boundary may lead to severe forgetting. 
This fundamental challenge, however, has not yet been studied extensively, especially in the setting where no samples from old classes are stored for rehearsal.
In this paper, we take a closer look at how the drift in the classifier leads to forgetting, and accordingly, design four simple yet (super-) effective solutions to alleviate the classifier drift: an \textbf{I}ndividual \textbf{C}lassifiers with Frozen Feature \textbf{E}xtractor (\textbf{\textsc{Ice}}) framework where we individually train a classifier for each learning session, and its three variants \textbf{\textsc{Ice}-\textsc{Pl}}, \textbf{\textsc{Ice}-\textsc{O}} and \textbf{\textsc{Ice}-\textsc{Pl\&O}} which further take the logits of previously learned classes from old sessions or a constant logit of an \textit{Other} class as constraint to the learning of new classifiers. Extensive experiments and analysis on 6 class-incremental information extraction tasks demonstrate that our solutions, especially \textbf{\textsc{Ice}-\textsc{O}}, consistently show significant improvement over the previous state-of-the-art approaches with up to 44.7\% absolute F-score gain, providing a strong baseline and insights for future research on class-incremental learning.\footnote{The source code, model checkpoints and data are publicly available at \url{https://github.com/VT-NLP/ICE}.}

\end{abstract}

\section{Introduction}


Conventional supervised learning assumes the data are independent and identically distributed (i.i.d.) and usually requires a pre-defined ontology, which may not be realistic in many applications in natural language processing (NLP). For instance, in event detection, the topics of interest may keep shifting over time (e.g., from \emph{attack} to \emph{pandemic}), and new event types and annotations could emerge incessantly.
Previous studies~\cite{ring1994continual, kirkpatrick2017overcoming, lopez2017gradient} therefore proposed continual learning (CL), a.k.a., lifelong learning or incremental learning, a learning paradigm aiming to train a model from a stream of \textit{learning sessions} that arrive sequentially. In this work, we focus on the \textit{class-incremental learning} (CIL) setting~\cite{wang2019sentence}, where a new \textit{session}\footnote{\textit{Session} is defined as an incremental learning stage to learn new classes with a model trained on the previous sessions.} is composed of previously unseen classes and the goal is to learn a unified model that performs well in all seen classes. 

\begin{figure}[tbp]
	\centering
	\includegraphics[width=0.9\linewidth, trim={0 0 0 0},clip ]{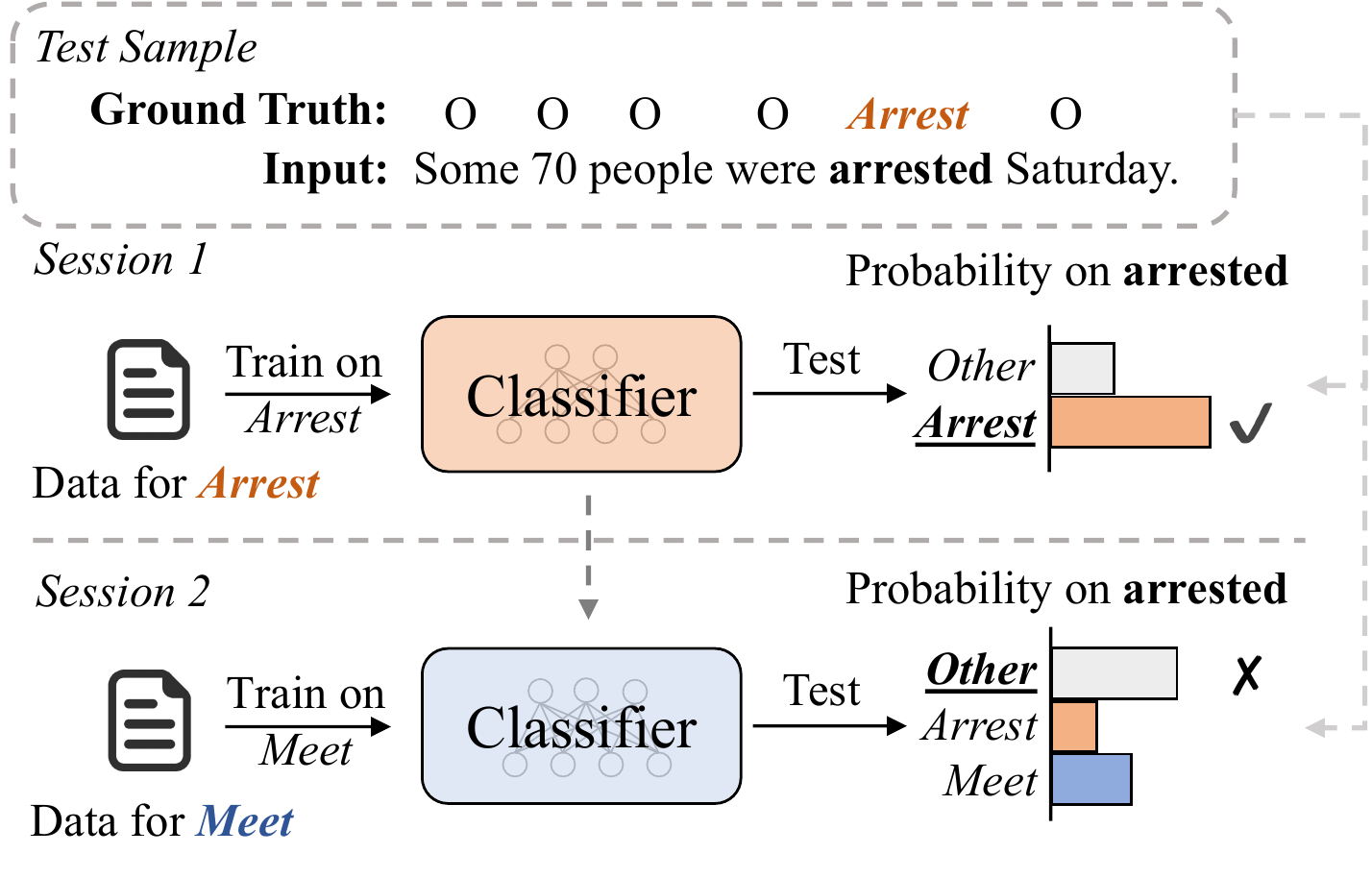}
 \vspace{-2mm}
	\caption{Illustration of class-incremental event detection where the model needs to classify each candidate mention into a label from all learned types or \textit{Other}. 
 The figure shows two classifiers that are incrementally trained from Session 1 and Session 2 and are evaluated on the same sample. After training on session 2, the classifier mistakenly predicts \textit{Other} for an \textit{Arrest} mention due to the \textit{classifier drift}.
 The model here uses pre-trained features and only the classifier is trained.
 }
	\label{fig_illustrate}
 \vspace{-5mm}
\end{figure}

When new learning sessions arrive sequentially, the classification layer must be constantly updated and/or expanded to accommodate new categories to the model. The change of the classifier between different sessions, i.e., \textit{classifier drift}, can disturb or overwrite the classifier trained on previous classes, which consequently causes catastrophic forgetting~\cite{biesialska-etal-2020-continual}. On the other hand, in many NLP tasks such as information extraction, the model also needs to classify negative instances into the \textit{Other} type (i.e., none-of-the-above). The \textit{Other} type adds extra difficulty to classification, and even worse, the meaning of \textit{Other} varies as the model learns new sessions~\cite{distillcausl2022zheng}. The CIL problem thus becomes even more challenging when \textit{Other} is involved. We illustrate the event detection task in CIL~\cite{yu2021lifelong} and the classifier drift problem in Figure~\ref{fig_illustrate}.


Despite the progress achieved in CIL~\cite{zhao2022consistent,distillcausl2022zheng}, 
there are two critical limitations that are still remained: (1) Most previous CIL approaches heavily rely on the rehearsal-based strategy which stores samples from previously learned sessions and keeps re-training the model on these examples in subsequent sessions to mitigate catastrophic forgetting, which requires high computation and storage costs and raises concerns about privacy and data leakage~\cite{privacy2015reza}; (2) Previous approaches have mainly focused on regularizing or expanding the overall model, especially feature extractor, to tackle the forgetting issue~\cite{cao2020increment}, but they rarely investigate whether the drift of the classifier also leads to forgetting, especially in classification tasks that involve the \textit{Other} category. In this work, we aim to tackle these limitations by answering the following two research questions: \textbf{RQ1}: \textit{how does classifier drift lead to forgetting in the setting where no samples are stored from old sessions for rehearsal?}, and \textbf{RQ2}: \textit{how to devise an effective strategy to alleviate classifier drift, especially when there is an Other category involved?}

In this paper, we aim to answer the two research questions above.
\textbf{First}, to study how classifier drift alone affects the model, we build a baseline where we use a pre-trained language model as a fixed feature extractor, such that only the parameters in the classification layer will be updated.
\textbf{Second}, to alleviate classifier drift, we propose a simple framework named \textbf{I}ndividual \textbf{C}lassifiers with Frozen Feature \textbf{E}xtractor (\textbf{\textsc{Ice}}). 
Instead of collectively tuning the whole classification layer, we individually train a classifier for the classes in each new session without updating old classifiers and combine all learned classifiers to classify all seen classes during inference. 
As individually trained classifiers may lack the context of all learned sessions ~\cite{envolved_cls}, they may not be comparable to each other. We further devise a variant \textbf{\textsc{Ice-Pl}} which takes the logits of previous classifiers as constraints to encourage contrastivity among all the classes when learning a new classifier for a new session.
\textbf{Third}, both \textbf{\textsc{Ice}} and \textbf{\textsc{Ice-Pl}} cannot be applied to detection tasks where an \textit{Other} class is involved, thus we further design two variants of them: \textbf{\textsc{Ice-O}} and \textbf{\textsc{Ice-Pl\&O}}, which introduce a constant logit for the \textit{Other} class and use it to enforce each individual classifier to be bounded by a constraint shared across different learning sessions during training. 

We extensively investigate the classifier drift and evaluate our approach on 6 essential information extraction tasks across 4 widely used benchmark datasets under the CIL setting. Our major findings and contributions are:
(1) By comparing the drifted baseline and our \textsc{Ice}, we find that the classifier drift alone can be a significant source of forgetting and our approaches effectively mitigate the drift and forgetting. Our results reveal that training the classifier individually can be a superior solution to training the classifier collectively in CIL.
(2) 
We find that the \textit{Other} type can effectively improve individually trained classifiers, and it is also helpful when we manually introduce negative instances during training on the tasks that do not have \textit{Other}.
(3) Experimental results demonstrate that our proposed approaches, especially \textbf{\textsc{Ice}-O}, significantly and consistently mitigate the forgetting problem without rehearsal and outperform the previous state-of-the-art approaches by a large margin. 
(4) Our study builds a benchmark for 6 class-incremental information extraction tasks and provides a super-strong baseline and insights for the following studies on class-incremental information extraction. 



\section{Related Work}



Existing approaches for CIL can be roughly categorized into three types~\cite{chen2022new}. 
Rehearsal-based approaches (a.k.a. experience replay)~\cite{lopez2017gradient,dautume2019episodic,guo2020improved,madotto-etal-2021-continual,lept5} select some previous examples (or generate pseudo examples) for rehearsal in subsequent tasks. While such approaches are effective in mitigating forgetting, they require high computation and storage costs and suffer from data leakage risk~\cite{privacy2015reza,memoryefficient,dualprompt}. 
Regularization-based approaches~\cite{llkd} aim to regularize the model's update by only updating a subset of parameters. Architecture-based approaches~\cite{Lee2020A,NEURIPS2021_bcd0049c,ke-etal-2021-classic,ke-etal-2021-adapting,feng2022hierarchical,zhu2022CPT} adaptively expand the model's capacity via parameter-efficient techniques (e.g., adapter, prompt) to accommodate more data. While most existing approaches consider alleviating the forgetting of the \emph{whole} model or transferring previous knowledge to new sessions,
few of them thoroughly investigate how the classification layer of the model is affected as it expands to incorporate more classes into the model.
\citet{bic} find that the classification layer has a strong bias towards new classes, but they only study this issue in image recognition that doesn't contain the \textit{Other} class.
To fill the blank in current research, we aim to take a closer look at how the drift in the classifier alone affects the model under the CIL setting, especially when \textit{Other} is involved.
For class-incremental information extraction, several studies tackle the CIL problem in relation learning~\cite{cml}, and many of them apply prototype-based approaches equipped with memory buffers to store previous samples~\cite{han2020continual,li2021refining,zhao2022consistent}. Others investigate how to detect named entities~\cite{extendNER,xia-etal-2022-learn} or event trigger~\cite{cao2020increment,yu2021lifelong,liu-etal-2022-incremental} in the CIL setting.
For instance, \citet{distillcausl2022zheng} propose to distillate causal effects from the \textit{Other} type in continual named entity recognition.
One critical disadvantage of existing approaches for continual IE is they heavily rely on storing previous examples to replay, whereas our method does not require any examplar rehearsal. 


\section{Problem Formulation}

Class-incremental learning requires a learning system to learn from a sequence of \textit{learning sessions} $\mathcal{D} = \{\mathcal{D}_1, ..., \mathcal{D}_T\}$ and each session $\mathcal{D}_k=\{(x^k,y^k)|y^k\in\mathcal{C}_k\}$ where $x^k$ is an input instance for the session $\mathcal{D}_k$ and $y^k\in\mathcal{C}_k$ denotes its label.
The label set $\mathcal{C}_k$ for session $\mathcal{D}_k$ is not overlapping with that of other sessions, i.e., $\forall k,j$ and $k\neq j, \mathcal{C}_k \bigcap \mathcal{C}_j=\emptyset$.
Given a test input $\mathbf{x}$ and a model that has been trained on up to $t$ sessions, the model needs to predict a label $\hat{\mathbf{y}}$ from a label space that contains all learned classes, i.e., $\mathcal{C}_1\bigcup...\bigcup\mathcal{C}_t$ and optionally the \textit{Other} class. 
Generally, the training instances in old classes are not available in future learning sessions. 

We consider a learning system consisting of a feature extractor and a classifier.
Specifically, we use a linear layer $\mathbf{G}_{1:t}\in\mathbb{R}^{c\times h}$ as the classification layer, where $c$ is the number of classes that the model has learned up to session $t$ and $h$ is the hidden dimension size of features. 
We denote the number of classes in a learning session $k$ as $n_k$, i.e., $n_k=|\mathcal{C}_k|$. The classification layer $\mathbf{G}_{1:t}$ can be viewed as a concatenation of the classifiers in all learned sessions, i.e., $\mathbf{G}_{1:t}=[\mathbf{W}_1;...;\mathbf{W}_t]$, where each of the classifier $\mathbf{W}_k\in\mathbb{R}^{n_k\times h}$ is in charge of the classes in $\mathcal{C}_k$.
The linear layer outputs the logits $o_{1:t}\in\mathbb{R}^{c}$ for learned classes, where $o_k$ refers to the logits for the classes in $\mathcal{C}_k$.
The term \textit{logit} we use in this paper refers to the raw scores \textit{before} applying the Softmax normalization. 

In this work, we focus on studying the class-incremental problem in information (entity, relation, and event) extraction tasks.
We consider two settings for each task: the \textit{detection} task that requires the model to identify and classify the candidate mentions or mention pairs into one of the target classes or \textit{Other}, 
and the \textit{classification} task that directly takes the identified mentions or mention pairs as input and classifies them into the target classes without considering \textit{Other}. 


\section{Approach}
\subsection{RQ1: How does Classifier Drift Lead to Forgetting?}
\label{sec_rq1}

We first design a \textsc{Drifted-Bert} baseline to investigate how classifier drift alone leads to forgetting, and then provide an insightful analysis of how classifier drift happens, especially in the setting of class-incremental continual learning.




\begin{figure*}[tbp]
	\centering
	\includegraphics[width=0.9\linewidth, trim={0 0 0 0},clip ]{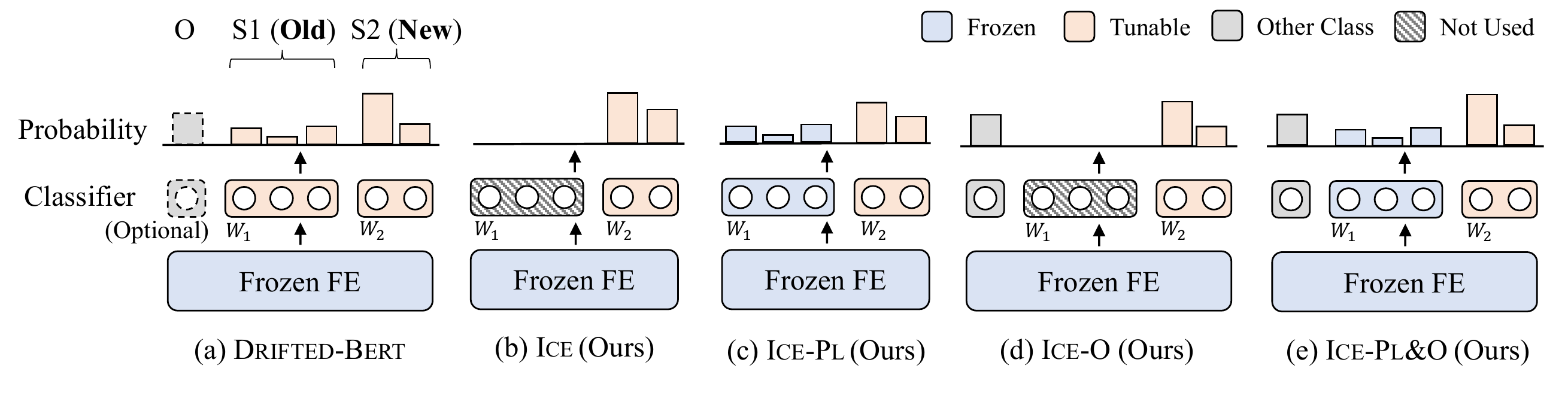}
 \vspace{-3mm}
	\caption{Illustration of the training process in a new learning session for \textsc{Drifted-Bert} as well as \textsc{Ice} and its variants. ``FE'' stands for \textit{feature extractor}. ``O'' stands for the \textit{Other} type. Each circle in the classifier represents a category. The models have learned a classifier ($\mathbf{W}_1$) with 3 classes in Session 1 (S1) and they are learning a new classifier ($\mathbf{W}_2$) with 2 classes in Session 2 (S2). \textsc{Ice} and \textsc{Ice-Pl} can only handle classification tasks without \textit{Other}, whereas \textsc{Ice-O} and \textsc{Ice-Pl\&O} are devised for detection tasks involving \textit{Other} where we use a fixed value as the logit of \textit{Other} class since it has distinct meanings in different sessions.  
 Note that \textsc{Drifted-Bert} is applied to both classification (w/o the \textit{Other} classifier) and detection (w/ the \textit{Other} classifier) tasks. 
 }
	\label{fig_main}
 \vspace{-5mm}
\end{figure*}

\paragraph{\textsc{Drifted-Bert} Baseline}

In the current dominant continual learning frameworks, both the feature extractor and classifier are continually updated, which results in drift in both components towards the model's predictions on old classes. To measure how the classifier drift along leads to forgetting, 
we build a simple baseline that consists of a pre-trained BERT~\cite{bert} as the feature extractor and a linear classification layer (shown in Figure~\ref{fig_main} (a)).
The model first encodes a given input text $x$ into the contextual representation. For event trigger and entity recognition, the model feeds the representation of a candidate span $\mathbf{h}$ into the linear layer to predict the logits for learned classes, i.e., $o_{1:t}=\mathbf{G}_{1:t}(\mathbf{h})$. For relation learning, we instead use the concatenation of head and tail representations as the feature, i.e., $\mathbf{h}=[\mathbf{h}_{head}; \mathbf{h}_{tail}]$.
For detection tasks, since each session contains an \textit{Other} class which has different meanings from other sessions, we follow~\cite{yu2021lifelong} to set the logit for \textit{Other} to a constant value $\delta$, i.e., $o_0 = \delta$. We combine $o_0$ and $o_{1:t}$ and pick the label with the maximum logit as the prediction.
That is, we predict a sample as \textit{Other} if and only if $max(o_{1:t})<\delta$.
We freeze the parameters in the feature extractor so that the encoded features of a given sample remain unchanged in different learning sessions.
In this way, the updates in the classification layer become the only source of forgetting. 
Note that we do not apply any continual learning techniques (e.g., experience replay) to \textsc{Drifted-Bert}. We denote $p(x^t)$ as the predicted probability to compute the loss in training, where $p(x^t)=\text{Softmax}(o_{0:t})$. At the learning session $t$, the model is trained on $\mathcal{D}_t$ with the Cross Entropy (CE) loss:
\begin{equation}
\mathcal{L}_{CE} = - \sum_{(x^t, y^t)\in\mathcal{D}_t} \log p(x^t).
\end{equation}

\paragraph{A Closer Look at Classifier Drift}

When the model has learned $t$ sessions and needs to extend to the $(t+1)$-th session, the classification layer $\mathbf{G}_{1:t}$ needs to introduce new parameters to accommodate the new classes in $\mathcal{C}_{t+1}$, i.e., $\mathbf{G}_{1:t+1}=[\mathbf{W}_1;...;\mathbf{W}_t; \mathbf{W}_{t+1}]$. As we assume that all previous training instances in $\mathcal{D}_{1:t}$ are not accessible anymore, solely training the model on $\mathcal{D}_{t+1}$ would lead to an extreme class-imbalance problem~\cite{cao2020increment}, which consequently causes catastrophic forgetting. However, most existing works rarely discuss how the drift in the classifier alone leads to forgetting, especially when the \textit{Other} class is involved. 

We first define the \textit{classifier drift} between two consecutive learning sessions $\mathcal{D}_t$ and $\mathcal{D}_{t+1}$ as \textit{the change from $\mathbf{G}_{1:t}$ to $\mathbf{G}_{1:t+1}$ that makes the model lose (part of) its acquired capability on the seen classes in $\mathcal{C}_{1:t}$.}
Intuitively, the CE loss aims to maximize the probability of the correct label while minimizing the probabilities of all other labels. Thus, there are two possible causes of classifier drift: (1) \textit{new logit explosion}: the new classifier $\mathbf{W}_{t+1}$ tends to predict logits $o_{t+1}$ that are higher than those of all previous classes $o_{1:t}$ so that the model can trivially discriminate new classes, which causes the old classes being overshadowed by new classes. (2) \textit{diminishing old logit}: as the old instances are not accessible in future learning sessions, the parameters in previous classifiers will be updated from the previous local optimum to a drifted sub-optimum, such that the classifier outputs low logits for old classes and cannot predict correctly. We empirically analyze the \textsc{Drifted-Bert} baseline to investigate the classifier drift in Section~\ref{sec_result_rq1} and discuss the drifting patterns in different classification and detection tasks in Section~\ref{app_logix_analysis}. 



\subsection{RQ2: How to Alleviate Classifier Drift?}
To alleviate the classifier drift, we introduce two solutions \textsc{Ice} and its variant \textsc{Ice-Pl} for the classification tasks without \textit{Other}, and further design two additional variants \textsc{Ice-O} and \textsc{Ice-Pl\&O} for detection tasks where \textit{Other} is involved. We illustrate the training process in a new learning session for \textsc{Ice} and its variants in Figure~\ref{fig_main}. Note that we only focus on the setting of continual learning without experience replay, i.e., the model does not have access to the data of old sessions. 

\paragraph{\textsc{Ice}: Individual Classifiers with Frozen Feature Extractor}


We revisit the idea of classifier ensemble~\cite{dietterich2000ensemble} and separated output layers in multi-task learning~\cite{zhang2018overview} where task-specific parameters for one task do not affect those for other tasks. Inspired by this, we propose to individually train a classifier for each session without updating or using previously learned classifiers $\mathbf{G}_{1:t}$ (shown in Figure~\ref{fig_main} (b)). In this way, previous classifiers can avoid being drifted to the sub-optimum, and the new classifier is less prone to output larger logits to overshadow old classes. 
Specifically, for an incoming session $t+1$, we initialize a set of new weights and train the new classifier $\mathbf{W}_{t+1}$ on $\mathcal{D}_{t+1}$. We only use the logits for the classes in the new session $o_{t+1}$ to compute the Cross-Entropy loss in optimization, i.e., $p(x^{t+1})=\text{Softmax}(o_{t+1})$.
During inference, as we need to classify all seen classes without knowing the session identity of each instance, we combine the logits from all classifiers $\mathbf{W}_{1},...,\mathbf{W}_{t+1}$ together to get the prediction for all learned classes, i.e., $o_{1:t+1}=[o_1;...;o_{t+1}]$, where each classifier yields the logits via $o_k=\mathbf{W}_{k}\cdot\mathbf{h}$ given the encoded feature $\mathbf{h}$ for each mention. 

\paragraph{\textsc{Ice}+Previous Logits (\textsc{Ice-Pl})} One limitation of \textsc{Ice} is the classifier individually trained in one session may not be comparable to others. To provide contrastivity to classifiers, we first explore a variant named \textsc{Ice-Pl} where we preserve the previous classifiers and only freeze their parameters, such that the new classifier is aware of previous classes during training (shown in Figure~\ref{fig_main} (c)). That is, the model uses the logits from all classifiers $o_{1:t+1}$ to compute the Cross-Entropy loss, i.e., $p(x^{t+1})=\text{Softmax}(o_{1:t+1})$, while only the parameters in the new classifier are trainable. \textsc{Ice-Pl} uses the same inference process as \textsc{Ice}.


\paragraph{\textsc{Ice}+Other (\textsc{Ice-O})}


Both \textsc{Ice-O} and \textsc{Ice-Pl} can only be applied to  classification tasks and handling the \textit{Other} category for detection tasks is challenging as each session $\mathcal{D}_t$ only contains the annotated mentions for the classes $\mathcal{C}_t$, while the mentions from all the other classes such as $\mathcal{C}_{1:t-1}$ are labeled as \textit{Other}, making the meaning of \textit{Other} varies in different sessions. To tackle this problem, we purpose the \textsc{Ice-O} variant (shown in Figure~\ref{fig_main} (d)) where we assign a constant value $\delta$ as the logit of the \textit{Other} category. 
Specifically, for each prediction, we combine the logit of \textit{Other} with the logits from the new session $o_{t+1}$ to obtain the output probability, i.e., $p(x^{t+1})=\text{Softmax}([\delta; o_{t+1}])$, and then compute the Cross-Entropy loss to train the classifier to make predictions for both positive classes and \textit{Other}. During the inference, we combine the \textit{Other}'s logit $\delta$ with the logits from all trained classifiers $o_{1:t+1}$, i.e., $o_{0:t+1}=[\delta;o_1;...;o_{t+1}]$ to predict for all learned positive types and \textit{Other}. We select the label with the highest logit among $o_{0:t+1}$ as the prediction, and a candidate will be predicted as \textit{Other} if and only if $max(o_{1:t+1})<\delta$. 

While the \textit{Other} class introduces additional difficulties to CIL, we argue that it can also be a good remedy to classifier drift. 
In particular, in each learning session $k$, while the classifier $\mathbf{W}_k$ is independently trained on $\mathcal{D}_k$, the output logits $o_k$ also need to satisfy the constraint that $max(o_k)<\delta$ when the classifier is trained on negative instances. Although the logits from any two distinct classifiers $\mathbf{W}_k$ and $\mathbf{W}_j$ ($k\neq j$) do not have explicit contrastivity, both classifiers are trained under the constraint that $max(o_k)<\delta$ and $max(o_j)<\delta$, which provides a weak contrastivity between them. 


\paragraph{\textsc{Ice}+Previous Logits and Other (\textsc{Ice-Pl\&O})} 
To explore the effect of preserving the previous logits when \textit{Other} is involved, we devise a \textsc{Ice-Pl\&O} variant that uses both the \textit{Other}'s logit $\delta$ and previous logits $o_{1:t}$ during training (shown in Figure~\ref{fig_main} (e)). That is, \textsc{Ice-Pl\&O} uses the combined logits $o_{0:t+1}=[\delta; o_1;...;o_{t+1}]$ to compute the loss, i.e., $p(x^{t+1})=\text{Softmax}(o_{0:t+1})$. \textsc{Ice-Pl\&O} adopts the same inference process as \textsc{Ice-O}.

While \textsc{Ice-O} and \textsc{Ice-Pl\&O} are naturally applied to detection tasks, for classification tasks without the \textit{Other} class, we can also manually create negative instances based on the tokens or entity pairs without positive labels.
Section~\ref{sec:data_exp_setup} provides more details regarding how to apply \textsc{Ice-O} and \textsc{Ice-Pl\&O} to classification tasks. 



\section{Experiments and Discussions}

\subsection{Datasets and Experiment Setup}
\label{sec:data_exp_setup}
We use \textbf{Few-NERD}~\cite{fewnerd} for class-incremental named entity recognition and split all the 66 fine-grained types into 8 learning sessions by following~\newcite{yu2021lifelong} which apply a greedy algorithm to split the types into sessions and ensure each session contains the roughly same number of training instances. We use two benchmark datasets \textbf{MAVEN}~\cite{maven} and \textbf{ACE-05}~\cite{ace} for class-incremental event trigger extraction and following the same setting as~\cite{yu2021lifelong} to split them into 5 learning sessions, respectively. For class-incremental relation extraction, we use \textbf{TACRED}~\cite{zhang2017tacred} and follow the same setting as~\newcite{zhao2022consistent} to split the 42 relations into 10 learning sessions.

For each dataset, we construct two settings: (1) \textit{detection} where the model classifies each token (or a candidate entity pair in relation extraction task) in a sentence into a particular class or \textit{Other}; and (2) \textit{classification} where the model directly takes in a positive candidate (i.e., an entity, trigger, or a pair of entities) and classify it into one of the classes. For the \textit{classification} setting, as there are no negative candidates that are labeled as \textit{Other}, we automatically create negative candidates and introduce the \textit{Other} category so that we can investigate the effect of \textit{Other} using \textsc{Ice-O} and \textsc{Ice-Pl\&O}.
Specifically, we assign the \textit{Other} label to the tokens if they are not labeled with any classes for entity and event trigger classification, and assign the \textit{Other} label to the pairs of entity mentions if they are not labeled with any relations for relation classification. When we apply \textsc{Ice-O} and \textsc{Ice-Pl\&O} to \textit{classification} tasks, during inference, we do not consider the logit of the \textit{Other} class.

\paragraph{Evaluation} We use the same evaluation protocol as previous studies~\cite{yu2021lifelong,liu-etal-2022-incremental}. Every time the model finishes the training on Session $t$, we evaluate the model on all test samples from Session 1 to Session $t$ for \textit{classification} tasks. For \textit{detection} tasks, we evaluate the model on the entire test set where we take the mentions or mention pairs of unlearned classes as \textit{Other}. Following~\newcite{yu2021lifelong}, we randomly sample 5 permutations of the orders of learning sessions and report the average performance.

\paragraph{Baselines}
We compare our approaches with the \textbf{\textsc{Drifted-Bert}} baseline and several state-of-the-art methods for class-incremental information extraction, including \textbf{ER}~\cite{wang2019sentence}, \textbf{KCN}~\cite{cao2020increment}, \textbf{KT}~\cite{yu2021lifelong}, \textbf{EMP}~\cite{liu-etal-2022-incremental}, \textbf{CRL}~\cite{zhao2022consistent}. All these methods adopt experience replay to alleviate catastrophic forgetting. We also design two approaches to show their performance in the conventional supervised learning setting where the model is trained with the annotated data from all the sessions, as the approximate upperbound of the continual learning approaches: (i) \textbf{BERT-FFE} consists of a pre-trained BERT as the feature extractor and a classifier, where, during training, we fix the feature extraction and only tune the classifier; and (ii) \textbf{BERT-FT} which shares the same architecture as \textbf{BERT-FFE} but both the feature extractor and classifier are tuned during training. More details about the datasets, baselines, and model implementation can be found in Appendix~\ref{app_exp_setup}.

\begin{table}[tbp]
	\centering
	
	\resizebox{0.49\textwidth}{!}
	{
		\begin{tabular}{l | c | ccccc}
    \toprule
    MAVEN (Detection) & Type & S1 & S2 & S3 & S4 & S5 \\
    
    \midrule
    \multirow{ 3}{*}{\textsc{Drifted-Bert}} & New & 50.9 & 57.8 & 52.8 &  52.7 & 49.1 \\
    & Acc-Old & - & 0 & 0 & 0 & 0 \\
    & Prev-Old & - & 0 & 0 & 0 & 0 \\
    \midrule
    \multirow{ 3}{*}{\textsc{Ice-O} (Ours)} & New & 50.9 & 56.0 & 53.2 & 49.9 & 49.3 \\
    & Acc-Old & - & \textbf{50.6} & \textbf{53.8} & \textbf{53.6} & \textbf{52.4} \\
    & Prev-Old & - & \underline{51.4} & \underline{56.2} & \underline{53.1} & \underline{50.0} \\
    \midrule
    \multirow{ 3}{*}{\textsc{Ice-O\&Pl} (Ours)} & New & 50.9 & 57.4 & 53.2 & 50.2 & 47.7 \\
    & Acc-Old & - & 50.3 & 53.0 & 52.7 & 50.7 \\
    & Prev-Old & - & 51.0 & 55.8 & 52.5 & 49.6 \\
    \bottomrule

    \midrule
    MAVEN (Classification) & Type & S1 & S2 & S3 & S4 & S5 \\
    \midrule
    \multirow{ 3}{*}{\textsc{Drifted-Bert}} & New & 86.9 & 63.1 & 54.7 & 47.6 & 34.0  \\
    & Acc-Old & - & 36.9 & 21.8 & 15.9 & 10.0 \\
    & Prev-Old & - & 36.4 & 33.4 & 29.4 & 29.1 \\
    \midrule
    \multirow{ 3}{*}{\textsc{Ice} (Ours)} & New & 86.9 & 79.8 & 72.8 & 68.0 & 59.2 \\
    & Acc-Old & - & 77.2 & 72.0 & 66.3 & 62.5 \\
    & Prev-Old & - & 77.5 & 72.1 & 65.7 & 62.6 \\
    \midrule
    \multirow{ 3}{*}{\textsc{Ice-Pl} (Ours)} & New & 86.9 & 67.5 & 57.2 & 49.2 & 34.9 \\
    & Acc-Old & - & 51.3 & 29.7 & 16.8 & 13.1 \\
    & Prev-Old & - & 51.1 & 49.5 & 37.2 & 38.5 \\
    \midrule
    \multirow{ 3}{*}{\textsc{Ice-O} (Ours)} & New & 86.5 & 79.8 & 76.9 & 73.3 & 63.8 \\
    & Acc-Old & - & 80.6 & \textbf{76.5} & \textbf{71.2} & \textbf{68.3} \\
    & Prev-Old & - & 81.0 & 76.1 & \underline{69.1} & \underline{69.4} \\
    \midrule
    \multirow{ 3}{*}{\textsc{Ice-Pl\&O} (Ours)} & New & 86.5 & 80.3 & 76.9 & 71.3 & 62.0 \\
    & Acc-Old & -  & \textbf{80.7} & 76.3 & 70.2 & 64.9 \\
    & Prev-Old & - & \underline{81.1} & \underline{77.0} & 67.9 & 66.2 \\

    \bottomrule
    \end{tabular}
}
	\caption{Analysis of the performance (Macro-F1 \%) on new and old classes on the class-incremental \textbf{event detection and classification} tasks on MAVEN. The best performance of accumulated old classes from all previous sessions (Acc-Old) is highlighted in \textbf{bold}, and the best performance of the old classes in the previous session (Prev-Old) is highlighted with \underline{underline}. 
 }
	\vspace{-1em}
	\label{tab_newold}
\end{table}



\begin{table*}[!htbp]
	\centering
	\resizebox{0.85\textwidth}{!}
	{
	\begin{tabular}{l | ccccc | ccccc | ccccc}
    \toprule
    & \multicolumn{5}{c}{MAVEN (Detection)}  & \multicolumn{5}{c}{ACE05 (Detection)} & \multicolumn{5}{c}{MAVEN (Classification)} \\
    \midrule
    Session &1 &2 & 3 & 4 & 5 & 1 &2 & 3 & 4 & 5 & 1 &2 & 3 & 4 & 5 \\ 
    \midrule
    ER\dag~\cite{wang2019sentence} & 62.0 & 48.6 & 43.1 & 35.5 & 32.5 & 59.6 & 46.2 & 41.7 & 33.2 & 34.4 & 88.9 & 70.4 & 64.1 & 55.0 & 53.1 \\
    KCN\dag~\cite{cao2020increment} & 63.5 & 51.1 & 46.8 & 38.7 & 38.5 & 58.3 & 54.7 & 52.8 & 44.9 & 41.1 & 88.8 & 68.7 & 59.2 & 48.1 & 42.1 \\
    KT\dag~\cite{yu2021lifelong} & \underline{63.5} & 52.3 & 47.2 & 39.5 & 39.3 & \underline{58.3} & 55.4 & 53.9 & 45.0 & 42.6 & 88.8 & 69.0 & 58.7 & 47.6 & 42.0 \\
    EMP\dag~\cite{liu-etal-2022-incremental} & \textbf{67.8} & \textbf{60.2} & 58.6 & 54.8 & 50.1 & \textbf{59.6} & 53.1 & 55.2 & 45.6 & 43.2 & \underline{91.5} & 54.2 & 36.7 & 27.0 & 24.8 \\
    CRL\dag~\cite{zhao2022consistent}  & - & - & - & -  & - & - & - & - & - & - & 89.2 & 73.2 & 70.0 & 63.7 & 62.9 \\
    \midrule
    \textsc{Drifted-Bert} & 60.5 & 41.0 & 33.8 & 22.5 & 20.8 & 53.7 & 50.6 & 51.8 & 20.1 & 17.2 & 90.1 & 52.3 & 39.7 & 28.0 & 22.3 \\
    \textsc{Ice} (Ours)  & - & - & - & - &  - & - & - & - & - &  -  & 89.4 & 79.0 & 75.8 & 71.4 & 68.5 \\
    \textsc{Ice-Pl} (Ours) & - & - & - & - &  - & - & - & - & - &  -  & 89.4 & 59.4 & 44.4 & 32.8 & 26.8 \\
    \textsc{Ice-O} (Ours) & 60.5 & \underline{59.9} & \textbf{61.3} & \textbf{60.8} & \textbf{61.4} & 53.7 & \underline{55.4} & \underline{60.7} & \underline{59.6} & \underline{61.5} & 88.8 & \underline{82.8} & \underline{81.0} & \underline{77.7} & \underline{75.5} \\
    \textsc{Ice-Pl\&O} (Ours) & 60.5 & 59.5 & \underline{60.7} & \underline{59.9} & \underline{60.2} & 53.7 & \textbf{55.8} & \textbf{61.4} & \textbf{60.5} & \textbf{62.4} & 88.8 & 82.1 & 79.8 & 75.2 & 71.6 \\
    \textsc{Ice-O}+TFE\&ER\dag (Ours) & 61.5 & 40.7 & 41.3 & 44.5 & 49.7 & 54.3 & 39.0 & 43.2 & 44.1 & 41.7 & \textbf{92.2} & \textbf{83.8} & \textbf{82.8} & \textbf{79.9} & \textbf{78.1} \\
    \midrule
    Upperbound (BERT-FFE) & - & - & - & - & 63.0 & - & - & - & - & 64.0 & - & - & - & - & 76.0 \\
    Upperbound (BERT-FT) & - & - & - & - & 67.3 & - & - & - & - & 66.6 & - & - & - & - & 81.0 \\
    \bottomrule
    \end{tabular}
	} 
 \vspace{-2mm}
	\caption{Results (Micro-F1 score, \%) on \textbf{event detection} and \textbf{classification} on 5 learning sessions. We highlight the best scores in \textbf{bold} and the second best with \underline{underline}. \dag\ indicates approaches with experience replay. 
 } 
	\vspace{-2mm}
	\label{tab_main_ed}
\end{table*}

\begin{table*}[htbp!]
	\centering
	\resizebox{0.9\textwidth}{!}
	{
	\begin{tabular}{l | cccccccc | cccccccc}
    \toprule
    & \multicolumn{8}{c}{Few-NERD (Detection)}  & \multicolumn{8}{c}{Few-NERD (Classification)} \\
    \midrule
    Session &1 &2 & 3 & 4 & 5 & 6 & 7 & 8 & 1 &2 & 3 & 4 & 5 & 6 & 7 & 8 \\ 
    \midrule
    ER\dag~\cite{wang2019sentence} & 57.7 & 45.7 & 45.5 & 39.7 & 34.1 & 28.3 & 23.7 & 23.5 & 93.7 & 60.6 & 49.9 & 42.4 & 38.4 & 32.3 & 30.0 & 26.0 \\
    KCN\dag~\cite{cao2020increment} & \underline{58.1} & 46.2 & 39.0 & 41.4 & 31.0 & 27.0 & 23.9 & 18.8 & 92.0 & 62.4 & 48.0 & 38.0 & 29.6 & 23.4 & 27.1 & 20.6 \\
    KT\dag~\cite{yu2021lifelong} & 57.7 & 46.8 & 44.0 & 42.5 & 35.9 & 26.0 & 24.9 & 23.3 & 93.3 & 58.7 & 51.2 & 41.1 & 34.3 & 25.6 & 20.9 & 20.2\\
    EMP\dag~\cite{liu-etal-2022-incremental} & \textbf{58.9} & 47.0 & 45.9 & 42.0 & 36.0 & 31.8 & 29.8 & 24.2 & \underline{94.0} & 52.0 & 39.7 & 32.0 & 26.3 & 22.0 & 24.6 & 17.6\\
    CRL\dag~\cite{zhao2022consistent} & - & - & - & - & - & - & - & - & 93.4 & 80.2 & 77.0 & 72.3 &  68.1 & 62.4 & 59.7 & 58.4 \\
    \midrule
    \textsc{Drifted-Bert} & 56.2 & 40.9 & 36.5 & 30.7 & 25.6 & 21.6 & 19.8 & 15.5 & 93.7 & 48.4 & 34.4 & 28.8 & 22.3 & 17.5 & 15.2 & 12.5 \\
    \textsc{Ice} (Ours) & - & - & - & - & - & - & - & - & 93.7 & 82.5 & 77.0 & 72.0 & 69.5 & 67.3 & 65.2 & 61.7 \\
    \textsc{Ice-Pl} (Ours) & - & - & - & - & - & - & - & - & 93.7 & 51.6 & 37.6 & 31.0 & 25.0 & 21.4 & 19.1 & 17.6 \\
    \textsc{Ice-O} (Ours) & 56.2 & \textbf{57.8} & \textbf{61.7} & \textbf{64.2} & \textbf{65.6} & \textbf{67.3} & \textbf{68.9} & \textbf{68.9} & 93.5 & \underline{86.6} & \underline{83.8} & \underline{80.4} & \underline{78.1} & \underline{76.5} & \underline{75.4} & \underline{71.9}  \\
    \textsc{Ice-Pl\&O} (Ours) & 56.2 & \underline{54.9} & \underline{57.1} & \underline{58.2} & \underline{58.9} & \underline{59.7} & \underline{60.6} & \underline{58.7} & 93.5 & 84.6 & 80.3 & 75.1 & 71.9 & 68.7 & 66.0 & 60.3 \\
    \textsc{Ice-O}+TFE\&ER\dag (Ours) & 50.7 & 42.2 & 45.0 & 45.4 & 46.2 & 48.7 & 47.5 & 47.1 & \textbf{94.2} & \textbf{87.7} & \textbf{86.5} & \textbf{83.9} & \textbf{82.0} & \textbf{81.7} & \textbf{80.2} & \textbf{76.1} \\
    \midrule
    Upperbound (BERT-FFE) & - & - & - & - & - & - & - & 72.3 & - & - & - & - & - & - & - & 73.5 \\
    Upperbound (BERT-FT) & - & - & - & - & - & - & - & 78.8 & - & - & - & - & - & - & - & 80.0 \\
    \bottomrule
    \end{tabular}
	} 
 \vspace{-2mm}
	\caption{
 Results (Micro-F1 score, \%) on \textbf{named entity recognition} and \textbf{classification} on 8 learning sessions. We highlight the best scores in \textbf{bold} and the second best with \underline{underline}. \dag\ indicates approaches with experience replay.
 } 
	\vspace{-4mm}
	\label{tab_main_ner}
\end{table*}




\subsection{RQ1: How does Classifier Drift Lead to Forgetting?}
\label{sec_result_rq1}

We conduct an empirical analysis on event detection and classification tasks on MAVEN to answer RQ1 and gain more insight into the classifier drift.

\paragraph{Analysis of Old and New Classes Performance}
Our first goal is to analyze the classifier drift during the incremental learning process. In Table~\ref{tab_newold}, we analyze 
how the performance of previously learned classes changes after the model has been trained on a new session for the \textsc{Drifted-Bert} baseline and the variants of \textsc{Ice}. 
After learning in each session $k$, we compute the (1) F-score on the new classes ($\mathcal{C}_k$) learned in the current session, (2) accumulated F-score on the old classes ($\mathcal{C}_{1:k-1}$) from all previous sessions, and (3) F-score on the old classes ($\mathcal{C}_{k-1}$) from the previous session, respectively. By comparing the performance change on the same set of classes in two continuous sessions, e.g., the F-score on the new classes ($\mathcal{C}_k$) learned in session $k$ and the F-score on the classes ($\mathcal{C}_k$) from the previous session after learning in session $k+1$, we can quantify how much the classifier is drifted. From Table~\ref{tab_newold}, the performance of \textsc{Drifted-Bert} on old classes after learning on a new session is always decreased dramatically, verifying that classifier drift does occur in class-incremental learning and leads to severe forgetting. On the other side, our solutions, especially \textsc{Ice-O}, consistently retain similar performance on the old classes from the previous session after learning on a new session, demonstrating that it effectively alleviates the classifier drift and the forgetting issue. 
Besides, we find that the \textsc{Ice-Pl} variant suffers from a considerable performance drop on both new and old classes, which indicates freezing previous classifiers' parameters while preserving the logits of previously learned classes cannot address the classifier drift and forgetting problems. 
Note that although we only showed the results on event classification and detection on MAVEN, the conclusions are very consistent for other tasks and datasets as shown in Appendix~\ref{app_more_old_new}. 

\subsection{RQ2: How to Alleviate Classifier Drift and Forgetting?}
\label{sec_result_rq2}

\begin{table*}[htbp!]
	\centering
	\resizebox{\textwidth}{!}
	{
	\begin{tabular}{l | cccccccccc | cccccccccc}
    \toprule
    & \multicolumn{10}{c}{TACRED (Detection)} & \multicolumn{10}{c}{TACRED (Classification)}   \\
    \midrule
    Session &1 &2 & 3 & 4 & 5 & 6 & 7 & 8 & 9 & 10 &1 &2 & 3 & 4 & 5 & 6 & 7 & 8 & 9 & 10 \\ 
    \midrule
    ER\dag~\cite{wang2019sentence} & 29.8 & \textbf{39.3} & \underline{34.1} & 31.8 & \underline{32.9} & 29.1 & 31.7 & 28.1 & 25.3 & 26.0 & 87.8 & 78.2 & 74.4 & 66.2 & 62.8 &  56.3 & 59.7 & 55.7 & 50.6 & 49.1 \\
    KCN\dag~\cite{cao2020increment} & 29.8 & 38.0 & 29.9 & 27.7 & 23.7 & 20.4 & 22.3 & 16.4 & 14.1 & 15.7 & 87.8 & 78.2 & 72.5 & 61.9 & 59.7 & 51.5 & 54.8 & 46.1 & 39.0 & 36.0 \\
    KT\dag~\cite{yu2021lifelong} & \underline{29.8} & 38.9 & 28.9 & 27.8 & 24.7 & 19.4 & 22.0 & 17.2 & 13.7 & 16.0 & 87.8 & 78.4 & 74.6 & 62.1 & 57.0 & 49.5 & 51.2 & 43.1 & 36.3 & 34.0 \\
    EMP\dag~\cite{liu-etal-2022-incremental} & 26.5 & \underline{39.1} & 31.8 & 30.3 & 30.1 & 23.8 & 31.3 & 23.8 & 21.3 & 21.1 & 88.0 & 54.2 & 44.5 & 37.4 & 32.4 & 29.9 & 35.6 & 33.8 & 21.1 & 27.5 \\
    CRL\dag~\cite{zhao2022consistent} & - & - & - & - & - & - & - & - & - & - & 88.7 & 82.2 & 79.8 & 74.7 & 73.3 & 71.5 & 69.0 & 66.2 & 64.0 & 62.8 \\
    \midrule
    \textsc{Drifted-Bert} & 28.9 & 36.7 & 27.7 & 26.5 & 21.8 & 17.4 & 21.2 & 17.6 & 13.7 & 14.3 & 88.8 & 51.0 & 30.9 & 27.0 & 17.2 & 17.8 & 19.0 & 14.7 & 10.8 & 14.3 \\
    \textsc{Ice} (Ours) & - & - & - & - & - & - & - & - & - & - & 88.8 & 77.8 & 73.4 & 67.5 & 60.7 & 55.6 & 56.8 & 52.6 & 51.1 & 49.2 \\
    \textsc{Ice-Pl} (Ours) & - & - & - & - & - & - & - & - & - & - & \underline{88.8} & 52.8 & 36.9 & 32.2 & 27.2 & 24.4 & 28.6 & 26.0 & 22.9 & 22.7  \\
    \textsc{Ice-O} (Ours) & 28.9 & 35.8 & \textbf{35.4} & \textbf{37.5} & \textbf{37.2} & \textbf{38.2} & \textbf{40.6} & \textbf{40.2} & \textbf{39.8} & \textbf{40.1} & 87.5 & \underline{85.7} & \underline{83.1} & \underline{81.4} & \underline{78.1} & \underline{75.8} & \underline{76.1} & \underline{72.0} & \underline{70.0} & \underline{67.4} \\
    \textsc{Ice-Pl\&O} (Ours) & 28.9 & 34.5 & 32.4 & \underline{33.0} & 30.3 & \underline{30.0} & \underline{32.0} & \underline{30.9} & \underline{29.5} & \underline{29.1} & 87.5 & 83.2 & 76.7 & 71.2 & 64.6 & 57.0 & 58.3 & 54.2 & 47.2 & 44.9 \\
    \textsc{Ice-O}+TFE\&ER\dag (Ours) & \textbf{33.4} & 13.2 & 12.6 & 14.8 & 16.4 & 18.8 & 22.4 & 24.5 & 26.1 & 27.7 & \textbf{95.2} & \textbf{92.1} & \textbf{91.2} & \textbf{90.8} & \textbf{88.6} & \textbf{86.1} & \textbf{86.3} & \textbf{83.6} & \textbf{82.7} & \textbf{81.4} \\
    \midrule
    Upperbound (BERT-FFE) & - & - & - & - & - & - & - & - & - & 51.2 & - & - & - & - & - & - & - & - & - & 73.3  \\
    Upperbound (BERT-FT) & - & - & - & - & - & - & - & - & - & 61.0 & - & - & - & - & - & - & - & - & - & 86.9  \\
   
    \bottomrule
    \end{tabular}
	} 
 \vspace{-1mm}
	\caption{
 Results (Micro-F1 score, \%) on \textbf{relation detection} and \textbf{classification} on 10 learning sessions. We highlight the best scores in \textbf{bold} and the second best with \underline{underline}. \dag\ indicates approaches with experience replay.
 } 
	\vspace{-1.2em}
	\label{tab_main_rc}
\end{table*}

To answer RQ2, we evaluate the effectiveness of our proposed approaches to mitigating classifier drift and catastrophic forgetting.

\paragraph{Quantitative Comparison}

We conduct an extensive quantitative comparison of the baselines and our approaches on the 6 class-incremental IE tasks. 
From Table~\ref{tab_main_ed},~\ref{tab_main_ner} and~\ref{tab_main_rc}, we can see that: 
(1) our approaches, especially \textsc{Ice-O}, without adopting experience replay, significantly and consistently alleviate the forgetting issue and show a remarkable improvement (i.e., ranging from 4.6\% - 44.7\% absolute F-score gain) over the previous state-of-the-art methods that are all based on experience replay. Notably, \textsc{Ice-O} achieves performance that is even close to the supervised \textbf{BERT-FFE} upperbound on most of the classification and detection tasks. 
(2) Among the four approaches, \textsc{Ice-O} consistently outperforms other variants on all the classification and detection tasks, demonstrating that 
introducing negative instances during training can constrain the updates in the classifier, and consequently mitigate classifier drift and forgetting.
(3) Persevering the logits of previous classes without updating the previous classifiers hurts the performance on most tasks, by comparing \textsc{Ice-Pl} with \textsc{Ice} and comparing \textsc{Ice-Pl\&O} with \textsc{Ice-O}. This observation is consistent with our findings in Section~\ref{sec_result_rq1}.  
(4) Previous methods generally perform worse than our solutions even with experience replay. The possible reasons include overfitting to the stored examples in the small memory buffer or the regularization from replay may not be effective enough to mitigate the forgetting.

\paragraph{Comparison with CRL~\cite{zhao2022consistent}} Note that, among all the baselines, \textbf{CRL} consistently outperforms others on the classification tasks. \textbf{CRL} is based on a prototypical network where each class is represented with a prototype computed from an embedding space and performs the classification with the nearest class mean (NCM) classifier. Compared with other Softmax-based classification approaches, \textbf{CRL} can accommodate new classes more flexibly without any change to the architecture. 
However, it still suffers from the \emph{semantic drift}~\cite{semantic_drift} problem as the embedding network must be continually updated to learn new classes, and it is non-trivial to adapt it to detection tasks where an \textit{Other} class is involved under the class-incremental learning setting and the meanings of \textit{Other} in different learning sessions are also different.

\paragraph{Comparison with Trainable Feature Extractor} We also investigate if our proposed approaches can be further improved by tuning the BERT-based feature extractor. However, it naturally leads to forgetting as demonstrated by previous studies~\cite{wang2019sentence,cao2020increment,yu2021lifelong}. Thus, following these studies, we adopt experience replay and design a new variant named \textsc{Ice-O} with Tunable Feature Extractor and Experience Replay (abbreviated as \textsc{Ice-O}+TFE\&ER), which tunes the BERT-based feature extractor and adopts the same replay strategy as \textbf{ER} that preserves 20 samples for each class. 
From Table~\ref{tab_main_ed},~\ref{tab_main_ner} and~\ref{tab_main_rc}, \textsc{Ice-O}+TFE\&ER significantly improves over \textsc{Ice-O}
and achieves comparable performance to the supervised \textbf{BERT-FT} upperbound on all the classification tasks. However, \textsc{Ice-O}+TFE\&ER performs much worse than \textsc{Ice-O} on all the detection tasks. We hypothesize that this is due to the meaning shift of the \textit{Other} class when incrementally training it on a sequence of learning sessions. Experience replay may not be enough to constrain the feature extractor to handle the \textit{Other} class properly.


\subsection{Analysis of Drifting Patterns}
\label{app_logix_analysis}
To take a closer look into how the classifier drift leads to forgetting and verify the two hypothetical drifting patterns we discuss in Section~\ref{sec_rq1}, we analyze the output logits (i.e., the scores before Softmax) from the old and new classifiers for \textsc{Drifted-Bert} and our \textsc{Ice}, \textsc{Ice-Pl}, and \textsc{Ice-O}. Specifically, we take the test samples whose ground truth labels are learned in Session 1 (denoted as $\mathcal{X}^{1}_{test}$), for analysis. Every time the classifier is trained on a new session, we evaluate the classifier on $\mathcal{X}^{1}_{test}$, and then take (1) the logit of the gold class (\textbf{Gold}), and (2) the maximum logit from the new classifier (\textbf{NCP}), i.e., New Classifier's Prediction, for analysis. For each type of logit, we report the average of the logits on all the samples in $\mathcal{X}^{1}_{test}$.

We have the following findings: 
(1) By examining the \textbf{Gold} logits and the logits from the new classifier (\textbf{NCP}) of \textsc{Drifted-Bert}, we observe that every time a new classifier is added and trained on the new session, the new classifier incrementally outputs higher logits than those in the previous session on $\mathcal{X}^{1}_{test}$ (\textcolor{fig_blue}{blue} solid line), whereas the \textbf{Gold} logits first decline a bit and stay at a certain level in the remaining sessions (\textcolor{fig_blue}{blue} dashed line). This observation confirms that two possible drifting patterns (i.e., \textit{new logit explosion} and \textit{diminishing old logit}) exist, and they can happen simultaneously and cause the new classifier overshadows the previously learned classifiers, which consequently leads to forgetting. 
(2) We find that while the old classifiers are not updated in \textsc{Ice-Pl}, the \textit{new logit explosion} issue gets even more severe (\textcolor{fig_orange}{orange} solid line), which explains why \textsc{Ice-Pl} performs worse than \textsc{Ice} and \textsc{Ice-O}. We hypothesize that the presence of previous logits may encourage the new classifier to predict larger logits. 
(3) When the classifier in each session is trained individually instead of collectively (i.e., in \textsc{Ice} and \textsc{Ice-O}), the \textbf{Gold} logits from the old classifiers stay at a constant level (\textcolor{fig_red}{red} dashed lines), whereas the logits from the new classifier are at a relatively lower level (\textcolor{fig_green}{green} and \textcolor{fig_red}{red} solid line). As such, the new classifier's logits do not have much impact on those of old classes, which mitigates the drift and forgetting. 

\begin{figure}[tbp]
	\centering
	\includegraphics[width=\linewidth, trim={0 0 0 0},clip ]{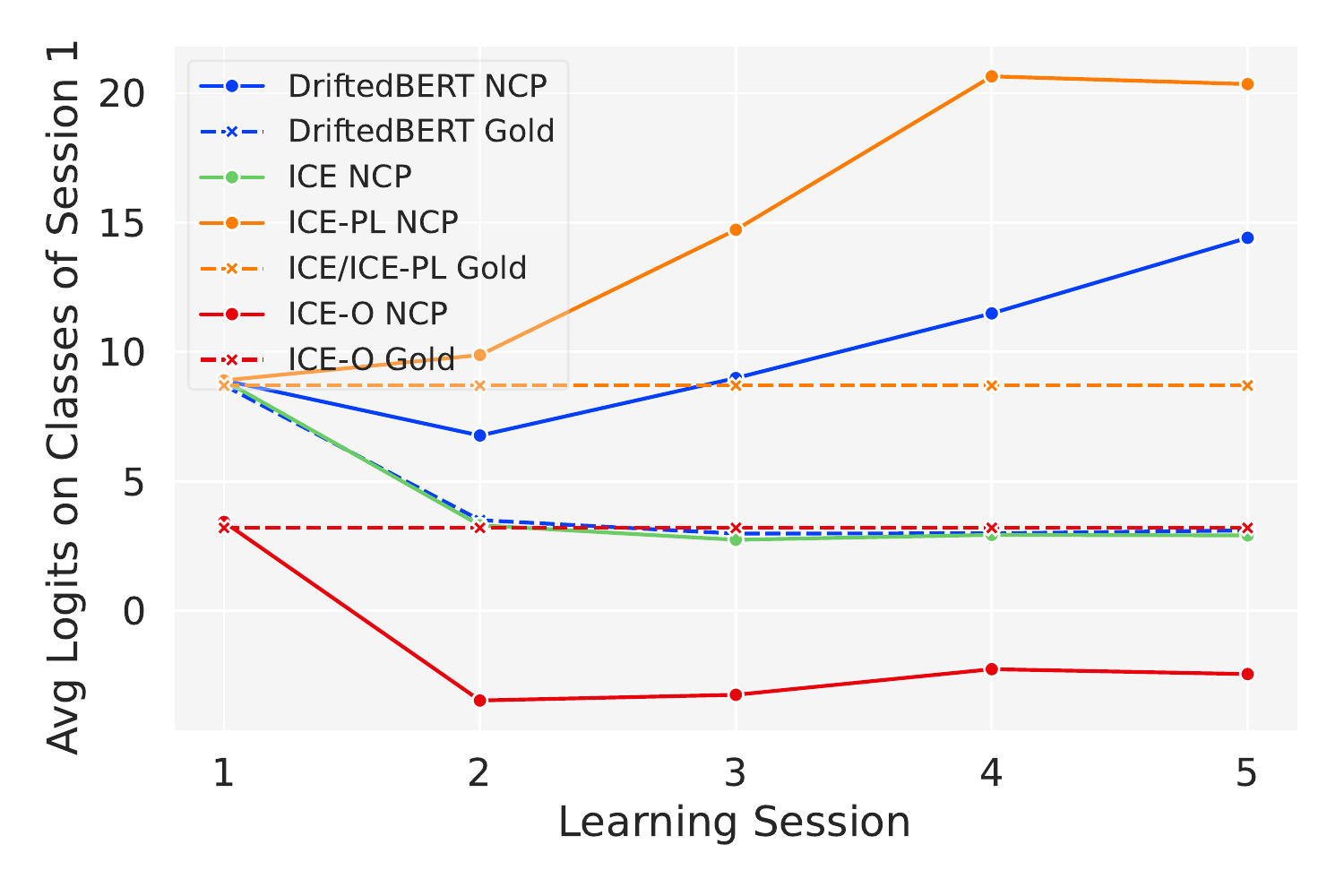}
	\caption{Analysis of output logits on the \textbf{event trigger classification} task on MAVEN. \textbf{Gold} refers to the gold logit and \textbf{NCP} refers to the maximum logit from the new classifier. We keep track of how these two types of logits change throughout 5 learning sessions.}
	\label{fig_logit}
\end{figure}

\subsection{The Effect of the Logit for \textit{Other} Class}
Throughout all the experiments, we set the logit for \textit{Other} class $\delta$ as 0 constantly. In this section, we further discuss the effect of the value of $\delta$, and the effect of tuning the \textit{Other} classifier. We show the results of event detection on MAVEN based on different fixed values or a tunable value of $\delta$ in Table~\ref{tab_other_logit}. We found that the value of \textit{Other} class's logit does not affect the model's performance much as long as it is fixed. However, we noticed a significant performance decrease if we continually tuned it with a classifier, demonstrating that it is necessary to fix the \textit{Other} class's logit during the continual learning process in our approach.

\begin{table}[tbp]
	\centering
	
	\resizebox{0.49\textwidth}{!}
	{
		\begin{tabular}{l | cccccccc}
    \toprule
    $\delta$ & 0 & 1 & 5 & 10 & -1 & -5 & -10 & Tune \\
    \midrule
    F1 & \textbf{61.6} & 61.5 & 61.6 & 61.0 & 61.4 & 60.8 & 60.2 & 56.5 \\
    
    \bottomrule
    \end{tabular}
}
	\caption{Results (Micro-F1 score, \%) on the effect of the \textit{Other} class's logit on the event detection task on MAVEN. We show the performance of the models that have learned all 5 sessions. “Tune” means we used a tunable logit for \textit{Other} class instead of a fixed value.
 }
	\vspace{-1em}
	\label{tab_other_logit}
\end{table}

\subsection{Comparison with Recent LLMs}
More recently, very large language models (LLMs) such as ChatGPT~\cite{ChatGPT} demonstrate strong in-context learning ability without the need of gradient update. Thus, class-incremental learning may also be tackled as a sequence of in-context learning. However, several recent studies~\cite{exploring2023gao,chatgpt2023qin} have benchmarked several LLMs with in-context few-shot learning on various IE tasks and show worse performance than our approach. 
Our approach can efficiently achieve a good performance that is close to the supervised performance by only finetuning the last linear layer using a much smaller frozen BERT backbone.
More critically, the knowledge LLMs are often bounded by the training data, whereas the goal of our continual learning approach focuses on incorporating up-to-date information into models.

\section{Conclusion}

In this paper, we investigate the answers and the solutions to the research questions that how the classifier drift alone affects a model in the class-incremental learning setting, and how to alleviate the drift without retraining the model on previous examples. We, therefore, propose to train a classifier individually for each task and combine them together during inference, such that we can maximally avoid the drift in the classifier. Extensive experiments show that our proposed approaches significantly outperform all the considered baselines on both class-incremental classification and detection benchmarks and provide super-strong baselines. 
We hope this work can shed light on future research on continual learning in broader research communities. 
\section*{Limitations}

Our approaches mainly leverage a fixed feature extractor together with a set of individually trained classifiers to mitigate catastrophic forgetting whereas a tunable feature extractor may also be helpful and complement the individually trained classifiers, so a future direction is to design advanced strategies to efficiently tune the feature extractor in combination with our proposed \textsc{Ice} based classifiers. In addition, we mainly investigate the classifier drift and demonstrate the effectiveness of our solutions under the class-incremental continual learning setting. Another future direction is to explore similar ideas under other continual learning settings, e.g., task-incremental learning, online learning, or the setting where new sessions also contain annotations for old classes.






\section*{Acknowledgments}
This research is based upon work partially supported by the Amazon Research Award program and U.S. DARPA KMASS Program \# HR001121S0034. The views and conclusions contained herein are those of the authors and should not be interpreted as necessarily representing the official policies, either expressed or implied, of DARPA or the U.S. Government. The U.S. Government is authorized to reproduce and distribute reprints for governmental purposes notwithstanding any copyright annotation therein.

\bibliography{custom}
\bibliographystyle{acl_natbib}

\appendix

\section{More Details on Experiment Setup}
\label{app_exp_setup}
\subsection{Details of the Datasets}
\label{app_dataset}
\paragraph{Named Entity} We use \textbf{Few-NERD}~\cite{fewnerd}, a large-scale named entity recognition (NER) dataset to evaluate class-incremental named entity recognition and classification. Compared with the datasets used in previous continual NER works~\cite{distillcausl2022zheng}, Few-NERD has a more diverse range of entity types and finer granularity, containing 8 coarse-grained and 66 fine-grained entity types. Thus, it is a better benchmark to study continual NER. 
We construct two settings for the NER task: (1) a detection task where the model is required to examine every token in the text and classify each of them into a learned positive entity type or \textit{Other}, and; (2) a classification task where the positive candidate entity mentions have been provided and the model only needs to assign a learned entity type to the given candidate.
Following~\newcite{yu2021lifelong}, we split the dataset into 8 learning sessions with the greedy algorithm such that each session contains the roughly same number of training instances. 

\paragraph{Relation} 
We use \textbf{TACRED}~\cite{zhang2017tacred} to evaluate relation detection and classification tasks. TACRED is a large-scale relation extraction dataset that contains 42 relations. In the previous continual relation classification setting~\cite{li2021refining}, they ignore the long-tail distribution and assume each relation contains the same number of instances. We instead use the original train/dev/test split in TACRED where relations are imbalanced. 
We build two settings for the relation task: (1) a detection task where the model needs to assign an ordered entity mentions with a seen positive relation type or \textit{Other}, and; (2) a classification task that assumes the given entity pair must belong to one of learned relation, and the model is only required to predict a label it has learned.
We follow the previous setting~\cite{zhao2022consistent} to split the dataset into 10 learning sessions, where we drop the relation with the fewest instances such that each session contains 4 positive relation types. 

\paragraph{Event Trigger} 
We adopt the following two event detection datasets for evaluation: 
(1) \textbf{MAVEN}~\cite{maven}: MAVEN is a large-scale event detection dataset with 169 event types (including \emph{Other}) in the general domain, and;
(2) \textbf{ACE-05}~\cite{ace}: ACE 2005 English dataset contains 34 event types (including \emph{Other}). For both datasets, we follow~\newcite{yu2021lifelong} to use the same train/dev/test split and use the same ontology partition to create 5 incremental learning sessions for each dataset, where each session contains approximately the same number of training instances.
We create two settings for event trigger: (1) two event detection tasks, where the model is required to evaluate each token in the sentence and assign it with a learned event type or \textit{Other}, and; (2) a classification task where the model only needs to classify a positive trigger mention into a learned event type without considering \textit{Other}.
We did not construct the classification task for the ACE dataset as the majority of instances only contain the \textit{Other} type and removing such instances will result in a very small dataset. 


\subsection{Baselines}
We use the following baselines for our experiments: 
(1) \textbf{\textsc{Drifted-Bert}}: we build a baseline with a fixed pre-trained BERT as the feature extractor and only train its classification layer. We do not apply any other continual learning techniques to it. We primarily use this baseline to study the \emph{classifier drift} discussed in this work.
(2) \textbf{ER}~\cite{wang2019sentence}: experience replay is introduced to continual IE by~\cite{wang2019sentence}. In this work, we use the same strategy as in~\cite{liu-etal-2022-incremental} to select examples to store in the memory and replay them in subsequent sessions. 
(3) \textbf{KCN}~\cite{cao2020increment}: the original work proposes a prototype-based method to sample examples to store for replay as well as a hierarchical knowledge distillation (KD) to constrain the model's update. We adapt their hierarchical distillation along with ER as the \textbf{KCN} baseline. 
(4) \textbf{KT}~\cite{yu2021lifelong}: a framework that transfers knowledge between new and old event types. 
(5) \textbf{EMP}~\cite{liu-etal-2022-incremental}: propose a prompt-based technique to dynamically expand the model architecture to incorporate more classes.
(6) \textbf{CRL}~\cite{zhao2022consistent} proposes consistent representation learning to keep the embeddings of historical relations consistent. Since \textbf{CRL} is designed for the classification tasks without \textit{Other}, we only evaluate this baseline on the classification tasks we build.
(7) \textbf{Upperbound}: we train a model jointly on all classes in the dataset as an upperbound in the conventional supervised learning setting. We devise two different upperbounds: (i) \textbf{BERT-FFE} is the upperbound of our \textsc{Ice-O} model, where we only train the classifier and the feature extractor is fixed. The negative instances are used in the classification tasks without \textit{Other}; and (ii) \textbf{BERT-FT} is the upperbound that trains both the whole BERT and the classifier.

\subsection{Implementation Details}
\label{exp_implement}
We use the pre-trained BERT-large-cased~\cite{bert} as the fixed feature extractor. We use AdamW~\cite{adamw} as the optimizer with the weight decay set to $1e-2$ and a learning rate of $1e-4$ for detection tasks and $5e-4$ for classification tasks. We apply gradient accumulation and set the step to 8. In each learning session $\mathcal{D}_k$, we establish a limit of 15 maximum training epochs. We also adopt the early stopping strategy with a patience of 3, where training will be halted if there is no improvement in performance on the development set for 3 epochs. We set the constant value for the \textit{Other} class $\delta$ to 0.
We apply the experience replay strategy with the same setting as in~\cite{liu-etal-2022-incremental} to \textbf{ER}, \textbf{KCN}, \textbf{KT}, and \textbf{EMP} as an assistant technique to mitigate forgetting. We store 20 examples for each class using the herding algorithm~\cite{herding} and replay one stored instance in each batch during training to limit the computational cost brought by rehearsal. For \textbf{CRL}, we use the same sample selection and replay strategy as in the original work. For baselines, we adopt a frozen pre-trained BERT-large and a trainable Multi-Layer Perceptron (MLP) as the feature extractor.

\section{More Discussions}

\subsection{More Analysis on Old and New Type Performance}
\label{app_more_old_new}
Table~\ref{tab_ner_old_new} and~\ref{tab_rel_old_new} show the performance of old and new classes for each learning session of the class-incremental named entity detection and classification and class-incremental relation detection and classification tasks.

\begin{table*}[htbp!]
	\centering
	\resizebox{\textwidth}{!}
	{
	\begin{tabular}{l | c | cccccccc | c | cccccccc}
    \toprule
    & \multicolumn{8}{c}{Few-NERD (Detection)}  & \multicolumn{8}{c}{Few-NERD (Classification)} \\
    \midrule
    Session & Type &1 &2 & 3 & 4 & 5 & 6 & 7 & 8  & Type & 1 &2 & 3 & 4 & 5 & 6 & 7 & 8 \\ 
    \midrule
    \multirow{ 3}{*}{ER\dag~\cite{wang2019sentence}} & New & 56.9 &  65.3 & 75.9 & 55.8 & 61.9 & 56.5 & 59.35 & 64.0 & New & 88.39& 74.2& 53.8& 42.1& 33.3& 25.0& 34.6& 26.0\\
    & Acc-Old & - & 34.4 &  11.0 & 18.1 & 20.3 & 21.0 & 23.9 & 19.1& Acc-Old & - & 50.6& 52.9& 33.7& 32.7& 34.2& 31.3& 33.9 \\
    & Prev-Old & - & 32.0 & 10.3 & 41.9 & 39.3 & 18.8 & 38.7 & 36.3 & Prev-Old & - & 48.9 & 58.1 & 48.8 & 51.0 & 37.3 & 42.3& 46.1\\
    \midrule
    \multirow{ 3}{*}{KCN\dag~\cite{cao2020increment}} & New &  64.3 & 58.6 & 57.9 & 61.3 & 56.3 & 76.0& 69.8 & 56.0& New & 88.3& 75.1& 63.1& 46.5& 33.4& 24.2& 31.1& 25.9 \\
    & Acc-Old & - & 33.0 & 35.5 &  18.6 & 12.8 & 6.5 & 6.5 & 11.9 & Acc-Old & - & 57.5& 52.4& 33.8& 29.4& 27.2& 25.5& 21.92 \\
    & Prev-Old & - & 24.44 & 39.8 & 18.6 & 17.9 & 11.9 & 25.0 & 44.6 & Prev-Old & - & 56.1 & 61.5 & 49.1 & 49.0 & 46.6 & 36.8 & 42.0\\
    \midrule
    \multirow{ 3}{*}{KT\dag~\cite{yu2021lifelong}} & New & 64.3 & 60.4 & 57.4 & 62.3 & 56.5 & 75.7 & 69.2& 58.2 & New & 88.3& 73.1& 60.6& 45.4& 34.7& 24.1& 34.0& 24.8\\
    & Acc-Old & - & 29.6 & 34.1 & 18.9 & 13.7 & 6.4 & 6.4& 8.6 & Acc-Old & - &51.2& 49.6& 27.6& 30.4& 27.6& 24.4& 26.7\\
    & Prev-Old & - & 17.3 & 39.8 & 21.9 & 25.0 & 13.8 & 22.0 & 35.0 & Prev-Old & - &49.3 & 58.8 & 45.8 & 48.8 & 34.7 &33.1 & 43.6\\
    \midrule
    \multirow{ 3}{*}{EMP\dag~\cite{liu-etal-2022-incremental}} & New & 61.9 & 56.6 & 53.1 & 58.4& 55.1 & 74.1 & 64.4 & 53.8 & New &  88.1& 75.4& 65.7& 49.1& 37.2& 31.7& 40.4& 23.1\\
    & Acc-Old & - & 37.2 & 41.0 & 36.9 & 32.3 & 18.7 & 25.0 & 26.5 & Acc-Old & - & 49.6& 56.5& 46.3& 44.6& 44.7& 39.7& 10.1\\
    & Prev-Old & - & 27.7 & 40.1 & 36.22 & 32.8 & 22.5 & 40.2 & 48.8 & Prev-Old & - & 46.7 & 70.1 & 58.6 & 56.0 & 47.9 & 47.1 & 15.5\\
    \midrule

    \multirow{ 3}{*}{\textsc{Drifted-Bert} } & New & 55.6 & 67.4 & 75.5 & 58.2 & 60.6 & 56.4 & 59.0& 59.4 & New & 88.1& 69.2& 60.5& 44.3& 32.0& 24.4& 36.8& 21.2\\
     & Acc-Old & - &5.5&  3.0 & 1.2 & 1.8 & 2.0& 1.4& 2.4& Acc-Old & - & 9.0& 4.4& 1.8& 10.7& 3.6& 6.4& 6.3\\
    & Prev-Old & - &0 & 0& 0& 0& 0& 0& 6.34& Prev-Old & - & 3.8 & 3.7 & 6.2 & 29.9 & 14.1 & 16.5 & 14.0\\
    \midrule
    \multirow{ 3}{*}{\textsc{Ice} (Ours)} & New & - & - & - & - & - & - & - & - & New & 88.1& 85.4& 85.5& 61.2& 63.7& 67.4& 65.4& 59.5\\
     & Acc-Old & -& - & - & - & - & - & - & - & Acc-Old & - & 79.9& 76.4& 72.0& 65.2& 63.7& 60.8& 59.0\\
    & Prev-Old & - & - & - & - & - & - & - & -& Prev-Old & - & 79.1 & 80.4 & 83.4 & 59.3 & 62.0 & 64.8 & 63.0\\
    \midrule
    \multirow{ 3}{*}{\textsc{Ice-Pl} (Ours) } & New & - & - & - & - & - & - & - & -& New & 88.1 & 70.9& 61.0& 45.0& 33.4& 26.2& 37.0& 21.2\\
     & Acc-Old & - & - & - & - & - & - & - & -& Acc-Old & - & 20.6& 6.7& 5.2& 16.1& 8.2& 9.9& 11.8\\
    & Prev-Old & - & - & - & - & - & - & - & -& Prev-Old & - & 16.1 & 9.2 & 36.6 & 46.4 & 36.0 & 42.5 & 41.5\\
    \midrule
    \multirow{ 3}{*}{\textsc{Ice-O} (Ours) } & New  & 55.6 & 69.0& 76.3& 61.0& 64.2& 62.7& 64.0& 68.8 & New & 87.8& 89.1& 89.3& 71.3& 74.9& 74.5& 72.2& 70.7 \\
     & Acc-Old & - & 58.2&  62.1& 62.9& 63.4& 64.4& 64.3& 63.8& Acc-Old & - & 82.5& 82.1& 78.5& 74.3& 72.9& 70.6& 67.9 \\
    & Prev-Old & - & 57.2 & 69.1 & 76.4 & 61.0 & 64.7 & 65.6 & 64.6& Prev-Old & - & 81.9 & 86.3 & 87.1 & 69.8 & 73.7 & 71.7 & 69.4 \\
    \midrule
    \multirow{ 3}{*}{\textsc{Ice-Pl\&O} (Ours)} & New & 55.6& 65.0& 73.3& 50.1& 57.1& 49.0& 51.4& 56.7 & New & 87.8& 87.9& 89.1& 66.5& 68.9& 61.3& 63.2& 58.9\\
     & Acc-Old & - &58.2& 60.2& 60.6& 57.4& 58.2& 56.4& 51.5& Acc-Old & - & 80.4& 80.1& 74.0& 67.4& 63.8 & 57.2& 53.2\\
    & Prev-Old & - &57.4 & 65.4& 74.2 & 51.0 & 59.5 & 56.3 & 50.3& Prev-Old & - & 79.7 & 84.9 & 85.0 & 65.1 & 68.1 & 59.2 & 60.2\\
   
    \bottomrule
    \end{tabular}
	} 
 \vspace{-2mm}
	\caption{
 Analysis of the performance (Macro-F1 \%) on new and old classes on the class-incremental \textbf{named entity detection and classification} tasks on Few-NERD.
 } 
	\vspace{-4mm}
	\label{tab_ner_old_new}
\end{table*}

\begin{table*}[htbp!]
	\centering
	\resizebox{\textwidth}{!}
	{
	\begin{tabular}{l | c | cccccccccc | c | cccccccccc}
    \toprule
    & \multicolumn{10}{c}{TACRED (Detection)}  & \multicolumn{10}{c}{TACRED (Classification)} \\
    \midrule
    Session & Type &1 &2 & 3 & 4 & 5 & 6 & 7 & 8 & 9 & 10  & Type & 1 &2 & 3 & 4 & 5 & 6 & 7 & 8 & 9 & 10 \\ 
    \midrule
    \multirow{ 3}{*}{ER\dag~\cite{wang2019sentence}} & New & 29.2 & 19.8& 23.2& 28.2& 27.1& 10.2& 47.5& 10.6& 36.6& 24.1   & New & 93.9& 53.9& 43.6& 50.9& 33.9& 22.2& 48.9& 21.6& 43.0& 27.3\\
    & Acc-Old & - & 32.9& 26.1& 16.6& 18.7& 19.7& 15.9& 17.3& 17.0& 17.5 & Acc-Old & - &69.5& 61.4& 49.3& 36.4& 41.9& 39.1& 34.9& 41.1& 40.1\\
    & Prev-Old  &- & 9.3 & 21.5 & 2.2 & 27.5 & 32.5 & 21.4 & 43.48 & 13.5 & 42.5& Prev-Old  &- & 63.8& 68.9& 54.6& 56.3& 49.5& 43.2& 61.3& 31.3& 66.4\\
    \midrule
    \multirow{ 3}{*}{KCN\dag~\cite{cao2020increment}} & New & 29.2& 19.8& 22.1& 28.3& 27.6& 8.2& 38.8& 11.3& 34.9& 21.7  & New & 95.3& 56.4& 40.1& 49.0& 33.9& 20.8& 45.0& 23.5& 49.0& 21.7\\
    & Acc-Old &- & 30.1& 15.9& 6.9& 10.8& 10.2& 13.2& 8.1& 7.6& 6.9 & Acc-Old & - &65.1& 49.4& 37.0& 35.7& 28.0& 30.0& 30.2& 32.6& 28.8\\
    & Prev-Old &- & 24.9 & 10.2& 0.6& 23.1& 34.8& 27.4& 27.5& 12.9& 40.5& Prev-Old & - &57.6& 43.7& 46.9& 58.6& 43.4& 32.0& 59.4& 32.9& 68.3\\
    \midrule
    \multirow{ 3}{*}{KT\dag~\cite{yu2021lifelong}} & New & 29.2& 19.6& 20.6& 26.0& 29.5& 10.1& 41.2& 11.1& 32.5& 24.5 & New &95.3& 60.8& 48.7& 50.6& 33.2& 16.2& 42.7& 18.3& 33.9& 22.0\\
    & Acc-Old &- & 30.2& 11.7& 10.0& 10.2& 9.6& 9.6& 6.8& 7.6& 7.9 & Acc-Old & - &58.4& 54.9& 47.4& 31.0& 24.7& 26.7& 27.2& 28.6& 21.6\\
    & Prev-Old &- & 25.3& 9.6& 12.1& 23.0& 34.5& 26.1& 24.9& 14.9& 41.6 & Prev-Old & - &58.4& 54.9& 47.4& 31.0& 24.7& 26.7& 27.2& 28.6& 21.6\\
    \midrule
    \multirow{ 3}{*}{EMP\dag~\cite{liu-etal-2022-incremental}} & New & 25.5& 19.3& 17.4& 32.3& 17.5& 9.5& 37.4& 8.4& 34.6& 19.3  & New & 90.9& 45.6& 35.6& 38.3& 29.3& 8.4& 40.8& 9.7& 17.7& 26.7\\
    & Acc-Old &- & 27.5& 19.9& 16.4& 11.7& 17.0& 17.0& 15.5& 12.8& 16.6& Acc-Old & - &44.5& 22.0& 19.4& 11.2& 15.4& 18.8& 8.2& 10.8& 18.2\\
    & Prev-Old &- & 22.9& 21.9& 5.9& 22.8& 31.9& 21.2& 34.1& 7.1& 45.8& Prev-Old & - &34.2& 25.8& 20.5& 11.7& 32.9& 3.7& 16.4& 5.5& 31.9\\
    \midrule

    \multirow{ 3}{*}{\textsc{Drifted-Bert} } & New & 28.7& 16.0& 19.3& 20.2& 21.8& 11.3& 43.2& 8.6& 40.7& 20.0 & New &93.6& 57.4& 35.6& 32.4& 29.9& 13.0& 36.3& 7.3& 18.3& 12.3\\
     & Acc-Old &- & 8.0& 3.9& 7.0& 4.6& 4.4& 1.8& 5.0& 4.7& 5.8& Acc-Old & - &18.6& 5.7& 4.6& 2.1& 4.6& 6.4& 3.1& 5.5& 6.9\\
    & Prev-Old  & - & 1.3& 1.8& 15.6& 12.4& 16.7& 0.0& 36.3& 9.0& 36.0& Prev-Old & - &0.0& 1.9& 6.3& 3.5& 22.5& 17.4& 13.8& 10.8& 47.1\\
    \midrule
    \multirow{ 3}{*}{\textsc{Ice} (Ours)} & New & - & - & - & - & - & - & - & - & - & - & New &93.6& 75.3& 33.6& 43.8& 47.2& 24.3& 49.8& 26.8& 58.1& 30.3\\
     & Acc-Old & - & - & - & - & - & - & - & - & - & -& Acc-Old & - &73.7& 55.6& 43.9& 36.1& 34.0& 31.2& 31.7& 30.2& 31.2\\
    & Prev-Old & - & - & - & - & - & - & - & - & - & -& Prev-Old & - &68.4& 56.3& 27.3& 45.1& 46.5& 20.3& 50.1& 26.8& 57.4\\
    \midrule
    \multirow{ 3}{*}{\textsc{Ice-Pl} (Ours) } & New & - & - & - & - & - & - & - & - & - & - & New &93.6& 57.4& 36.7& 33.3& 27.7& 16.2& 37.0& 15.9& 61.8& 23.8\\
     & Acc-Old & - & - & - & - & - & - & - & - & - & -& Acc-Old & - &18.6& 7.2& 9.3& 9.1& 9.5& 8.2& 10.7& 13.2& 15.9\\
    & Prev-Old & - & - & - & - & - & - & - & - & - & - & Prev-Old & - &0.0& 4.9& 21.1& 33.3& 46.9& 28.9& 39.0& 16.8& 64.9\\
    \midrule
    \multirow{ 3}{*}{\textsc{Ice-O} (Ours) } & New & 28.7& 16.4& 19.3& 28.5& 29.4& 20.2& 42.2& 7.3& 38.2& 26.1  & New &92.7& 78.0& 45.6& 61.3& 61.0& 34.5& 64.2& 33.5& 70.0& 33.8\\
     & Acc-Old &- &31.0& 23.0& 23.1& 24.5& 26.5& 26.3& 26.5& 26.2& 26.4& Acc-Old & - &86.4& 71.3& 60.4& 58.0& 55.0& 50.8& 48.0& 46.8& 45.0\\
    & Prev-Old &- & 29.9& 16.2& 19.8& 29.3& 30.6& 22.3& 42.2& 7.4& 38.1& Prev-Old & - &84.2& 73.6& 44.7& 61.1& 54.5& 30.0& 63.9& 33.2& 69.9\\
    \midrule
    \multirow{ 3}{*}{\textsc{Ice-Pl\&O} (Ours)} & New  & 28.7& 17.2& 18.1& 20.3& 16.5& 7.7& 35.1& 6.4& 34.8& 15.2 & New & 92.7& 67.9& 50.0& 55.8& 43.5& 19.8& 53.8& 24.8& 57.0& 21.4\\
     & Acc-Old &- & 33.1& 23.4& 22.3& 18.7& 19.4& 18.9& 19.7& 20.4& 21.6& Acc-Old & - &81.5& 62.0& 50.4& 38.9& 31.1& 28.1& 29.2& 27.2& 26.5\\
    & Prev-Old &- & 30.5& 17.5& 19.8& 22.2& 18.1& 8.2& 33.9& 7.2& 36.4& Prev-Old & - &78.6& 67.2& 49.1& 46.3& 50.9& 21.5& 53.2& 24.2& 61.0\\
   
    \bottomrule
    \end{tabular}
	} 
 \vspace{-2mm}
	\caption{
 Analysis of the performance (Macro-F1 \%) on new and old classes on the class-incremental \textbf{relation detection and classification} tasks on TACRED.
 } 
	\vspace{-4mm}
	\label{tab_rel_old_new}
\end{table*}

\end{document}